\documentclass[10pt,final,journal,twocolumn]{IEEEtran}
\usepackage[utf8]{inputenc}
\usepackage{cite}
\usepackage[cmex10]{amsmath}
\usepackage{amsmath}
\usepackage{algorithm}
\usepackage{multirow}
\usepackage{multicol}
\usepackage{stfloats}
\usepackage{multicol,multienum}
\usepackage{multirow}
\usepackage[caption=false,font=footnotesize]{subfig}
\usepackage{wrapfig}
\usepackage{color}
\usepackage{graphicx}
\usepackage{xspace}

\def\ie{\textit{i.e.}\xspace}

\def\etc{\textit{etc.}\xspace}

\def\eg{\textit{e.g.}\xspace}

\begin{document}

\title{Green Deep Reinforcement Learning for Radio Resource Management: Architecture, Algorithm Compression and Challenge
\thanks{Zhiyong Du is with National University of Defense Technology, Changsha, China. Yansha Deng is with Kings College London, London, United Kingdom. Weisi Guo is with Cranfield University, Bedford, United Kingdom and Alan Turing Institute, London, United Kingdom. Arumugam Nallanathan is with Queen Mary University of London, London, United Kingdom. Qihui Wu is with Nanjing University of Aeronautics and Astronautics, Nanjing, China.}
\thanks{This paper is partly funded by EC H2020 grant 778305 and NSF of China under Grants 61601490. $^*$Corresponding Author: wguo@turing.ac.uk}
}

\author{
\IEEEauthorblockN{Zhiyong Du, Yansha Deng, Weisi Guo, Arumugam Nallanathan \textit{IEEE Fellow}, Qihui Wu\\}}

\maketitle

\begin{abstract}
AI heralds a step-change in the performance and capability of wireless networks and other critical infrastructures. However, it may also cause irreversible environmental damage due to their high energy consumption. Here, we address this challenge in the context of 5G and beyond, where there is a complexity explosion in radio resource management (RRM). On the one hand, deep reinforcement learning (DRL) provides a powerful tool for scalable optimization for high dimensional RRM problems in a dynamic environment. On the other hand, DRL algorithms consume a high amount of energy over time and risk compromising progress made in green radio research. 

This paper reviews and analyzes how to achieve green DRL for RRM via both architecture and algorithm innovations. Architecturally, a ``cloud based training and distributed decision-making'' DRL scheme is proposed, where RRM entities can make lightweight deep local decisions whilst assisted by on-cloud training and updating. On the algorithm level, compression approaches are introduced for both deep neural networks and the underlying Markov Decision Processes, enabling accurate low-dimensional representations of challenges. To scale learning across geographic areas, a spatial transfer learning scheme is proposed to further promote the learning efficiency of distributed DRL entities by exploiting the traffic demand correlations. Together, our proposed architecture and algorithms provide a vision for green and on-demand DRL capability. 
\end{abstract}


\IEEEpeerreviewmaketitle
\section{Introduction}
Future AI driven automation of wireless networks and other critical infrastructures will bring about a step change in their ability to create efficient, resilient, and also user-centric services. However, the very same algorithms may also cause irreversible environmental damage due to their high energy consumption and lead to serious global sustainability issues. 

The wireless ICT industry is one of the fastest growing carbon emission industries, and will millions of base stations and billions of smart phones deployed worldwide. To meet the rapidly increasing traffic volume and demand diversity across network slices, 5G and beyond mobile networks are expected to introduce a number of fundamental innovations apart from PHY and MAC layer technology enhancements. To allow for centralized and large-scale network coordinated optimization, software defined network (SDN) and network function visualization (NFV) have been proposed to optimize both the radio access network and mobile core network by integrating data analytics and cloud control. On the other hand, vertical industries, such as manufacturing, automotive and health-care, \etc, will impose diverse performance and experience requirements on the latency, throughput, and reliability. This brings the need to evolve beyond cognitive radio towards an artificially intelligent (AI) resource optimization ecosystem to support more fine-grained user-centric service provision (see 3GPP Release 16 TR37.816). This becomes more challenging in highly dynamic environments involving UAV 3D heterogeneous channels. As a result, radio resource management (RRM) is becoming increasingly complex and high dimensional parameter optimization could be a concern.

The growing complexity problem in wireless ecosystems cannot be solved in a scalable manner by the traditional optimization approaches, such as dynamic programming, convex optimization, \etc, as they predominantly work on the premise that a known optimization model is available. Recently, the success of deep reinforcement learning (DRL) has opened new pathways to scalable optimization for high dimensional problems in wireless communications and networks. DRL retains the model-free optimization capability of traditional reinforcement learning (RL), suitable for dynamic and online RRM. Meanwhile, in DRL, deep neural network (DNN) is used to approximate policy or value functions for large-scale RL problem, overcoming the intrinsic scalability issue of traditional tabular RL approaches. Specifically, the powerful function approximation and representation learning properties \cite{DRL_survey} of DNN empower RL with robust and high efficient learning. The application of DRL in 5G and beyond\cite{DRL_application_survey} shows great promise and is receiving more and more attention in the community at both the PHY and MAC layers\footnote{see IEEE ComSoc Best Reading: https://www.comsoc.org/publications/best-readings/machine-learning-communications}. 

Most existing RRM solutions applied in RRM use off-the-shelf algorithms with little consideration on the RRM feature set. Different from supervised DL applications, where a large amount of samples are available in advance for training, the training samples in DRL can be only generated from the interaction between the RL agent and the wireless network environment. In particular, since one interaction iteration in RRM commonly involves parameter configuration and feedback acquisition, the time penalty is not negligible. 

A growing concern in the machine learning community is the high energy consumption in DRL. A common DNN consists of several stacked layers of neurons with tens or up to hundreds of millions of weights. Such a large number of parameters will generate high computational burden and memory access processes during both training and inference stages. Even sufficient computation capability is provided, the resulting energy consumption is unacceptable for reinforcement learning, especially in battery-constrained devices and areas that do not have access to green electricity supply. For example, smartphones nowadays cannot even run object classification with AlexNet in real-time for more than an hour \cite{Energy-Efficient}. 

In view of these challenges, this paper studies how to achieve green DRL for RRM. First, we review the state-of-the-art in Section II. In Section III, we envision an efficient DRL architecture for RRM and outline both architectural design improvements and algorithmic methodological advances. To provide RRM entities with affordable and on-demand DRL capability, a ``cloud based training and distributed decision-making'' architecture is proposed, balancing distributed decision-making with cloud based training and updating. In Section IV, to reduce the computation and energy consumption in DRL algorithms, several algorithm compression approaches are introduced, including deep neural network compression, MDP model compression and spatial transfer learning scheme. Finally, some challenges are analyzed in Section V.

\section{State of the Art}
\label{section2}

\subsection{RRM Formulation}
To apply DRL to RRM, it is necessary to map the considered problem into an appropriate DRL model. The core elements of DRL are: (i) state, (ii) action, (iii) reward, (iv) a model of environment dynamics, (v) policy and (vii) DNN implementation \cite{DRL_survey}. The former four elements define the underling Markov Decision Process (MDP) of RL. Specifically, a learner or agent interacts with environment at discrete time epochs $t=0, 1, 2, ...$ as follows:
\begin{itemize}
    \item Each time the agent observes some representation $\boldsymbol{s}_t\in \mathcal{S}$ of the environment state, where $\mathcal{S}$ is the set of states, it selects an action $\boldsymbol{a}_t$ from the action set $\mathcal{A}$;
    \item At the beginning of the next time epoch $t+1$, the agent receives a delayed numerical reward $r_t$ and the environment state evolves to state $\boldsymbol{s}_{t+1}$; 
\end{itemize}
The \textit{goal} is to find a policy $\pi(\boldsymbol{s},\boldsymbol{a})$ that maps states to probabilities of selecting each possible action in order to maximize the expected sum of the discounted rewards $E_{\pi}\left[ \sum_{t=0}^{\infty}{\lambda^t r_t} \right]$. The parameter $E_{\pi}$ is the expectation under policy $\pi$, and $0< \lambda \leq 1$ is a discount factor. In absence of the information on the environment dynamics, \ie, $p\left( \boldsymbol{s}'\mid \boldsymbol{s},\boldsymbol{a} \right) $, $\boldsymbol{s}'$, $\boldsymbol{s}\in \mathcal{S}$, which is common in wireless network applications, traditional dynamic programming is intractable in solving MDP. Alternately, RL algorithms are proposed to learn the optimal policy from interaction without a model of the environment dynamics. The DNN is used to approximate the optimal policy or value function for large-scale RL problems.

To support the ambitious goal of 5G and beyond mobile networks, the general RRM problem can be seen as ``\textit{realizing context-aware optimization to maximize expected accumulative target KPI such as system QoS or user QoE}''. From this perspective, the mapping between DRL and RRM optimization can be constructed as follows. The communication contexts that specify the situation of user and networks corresponds to states, which may include user profile (location, user demand, mobility pattern, \etc), spectrum environment (spectrum usage, interference), link state (wireless channel gain, data rate, reliability), network state (traffic distribution, packet buffer of nodes, available slice resource, power and energy consumption, \etc), application information (content cache) and configuration parameters. The target configuration parameters in the considered problem are the actions. The user QoE and/or system KPI achieved in each decision epoch is the instant reward. In some scenarios, it is impractical to observe complete state information. For example, the instant quality state of all channels or traffic generation dynamics of all terminals are hard to acquire. This results in a more challenging RL problem under the partially observable Markov decision process (POMDP) model.

\subsection{DRL Design}
Different variations of DRL algorithms have been introduced for RRM, where the RL mechanism and the architecture of DNN largely determine the characteristics of DRL. 

\subsubsection{Reinforcement Learning Mechanism}
The RL mechanisms in DRL can be broadly classified into three types \cite{DRL_survey}: value function based method, policy based method, and actor-critic method. 

In \textit{value function} based methods, the action-value function $Q^{\pi}\left( \boldsymbol{s,a} \right)$ defined by the long-term return when starting in state $\boldsymbol{s}$ with action $\boldsymbol{a}$ and following policy $\pi$ subsequently, are estimated and the optimal action(s) for each state correspond to the one(s) with the largest action-value. In \textit{policy based} DRL, a DNN is used to directly derive the  stochastic policy, \ie, mapping input state vector to selection probability distribution over all  actions. Searching the optimal policy can be gradient-free or gradient-based methods. Note that one advantage of policy based method is that it can handle continuous action, which may be preferred for possible continuous parameters, such as power control, location and distance optimization. In this context, the output are the mean and standard deviations of Gaussian distribution. Finally, the \textit{actor-critic} method is a combination of the above two methods: the state value function is introduced to generate feedback for policy gradient. There are different deep actor-critic algorithm variations, one of most widely used is the asynchronous advantage actor-critic (A3C) that could even run on parallel and asynchronously. 

\begin{table*}[!t]
\renewcommand{\arraystretch}{1}
\caption{Related work in using DRL for resource management.}
\label{table1}
\centering
        \begin{tabular}{|l|l|l|l|l|}
        \hline
       \textbf{Related Work} & \textbf{State}& \textbf{Action}& \textbf{DNN}& \textbf{RL Mechanism}\\
       \hline
               \multirow{2}{*}{\textbf{Random Access  Control \cite{RL-IoT-MassC-JSAC}}}
               & historical number of  idle, collided   &{number  of allocated RACH, }& \multirow{2}{*}{full connection} &\multirow{2}{*}{value function}\\
       & and successful channels& preambles and repetition values &  &\\
       \hline
       \multirow{2}{*}{\textbf{Power Control \cite{DRL_power, DRL-Power-Access}}} &require load and interference values; &power allocation (continuous);& convolution layer;&policy based;\\
       &RSS measurements &power allocation (discrete) &full connection & value function\\
       \hline
               \multirow{2}{*}{\textbf{Spectrum Sharing \cite{DRL-Coverage-TWC, DRL-Coverage-TCCN, DRL-HetNet}}}
               &selected channel, capacity and ACK;&\multirow{2}{*}{selected channel}& recursion connection;&\multirow{2}{*}{value function}\\
       & selected channels and conditions& & full connection &\\
      \hline
              \multirow{2}{*}{\textbf{Coverage \& Connectivity \cite{DRL_RAN}}} &active/sleep state and traffic &\multirow{2}{*}{on/sleep of radio head}& \multirow{2}{*}{full connection }&\multirow{2}{*}{value function}\\
       &demand of radio heads& & &\\
       \hline
              \multirow{2}{*}{\textbf{Mobility Management \cite{DRL_handover}}}
                &\multirow{2}{*}{RSRQs of all cells and serving cell}&\multirow{2}{*}{selected cell}& \multirow{2}{*}{recursion connection}  & \multirow{2}{*}{actor-critic}\\
                & &    & &\\
        \hline
        \multirow{2}{*}{\textbf{Slice Management\cite{ DRL_slice, DRL_slice2}}}
        &number of arriving packets in slices;&allocated bandwidth to slices;& \multirow{2}{*}{full connection}  &\multirow{2}{*}{value function}\\
        &number of slices of each class         &   access control of slices   & &\\
          \hline
        \end{tabular}
\end{table*}

\subsubsection{Neural Network Architecture and Relation to RRM Problems}
DNNs are used to exploit the potential correlation of states, actions and policies for efficient approximation of high-dimensional RL problem. Three typical deep neural network (DNN) architectures are widely used. 

The first type is \textit{full connection}, where each neuron in hidden layers is connected with all neurons in the previous and next layers. It is a deep version of traditional multi-layer perceptron  and is commonly used in combination with feature detection layers. 

The second type is \textit{convolution layer}. A neuron in a convolution layer is connected to local patches in the feature maps of the previous layer sharing the same weight. The operation of convolution can be seen as a filter to local groups, which is suitable for array data processing and has the advantage of detecting spatial correlation of states in resource management problems. For example, exploring geography-dependent correlated shadowing, channel gains or complex spectrum interference patterns to enhance spectrum allocation, and extracting temporal-spatial traffic demand distribution for on-demand schedule. Instead of using complex convolutional neural network (CNN), the application of convolution architecture in DRL is relatively flexible, \ie, it is only used as a lightweight feature extraction layer. 

The third type is \textit{recursive connection} specialized for processing sequential data. The recursion means neurons also take other neurons' outputs at previous time steps as inputs, to store history information for predicting future output of the sequential data. The popular recurrent neural network (RNN) can predict user behaviors in wireless communication. For example, the content or application request data can be trained to predict service request to achieve personalized service provisioning. In addition, RNN can predict user mobility pattern to improve small cell handover in mobility management. 

There are also some RNN variations with memory network architecture that is suitable for longer memory requirement cases, such as long short term memory (LSTM) network. 

\vspace*{-0.0cm}
\begin{figure}[htbp!]
    \begin{center}
    \centering
        \includegraphics[width=0.5\textwidth]{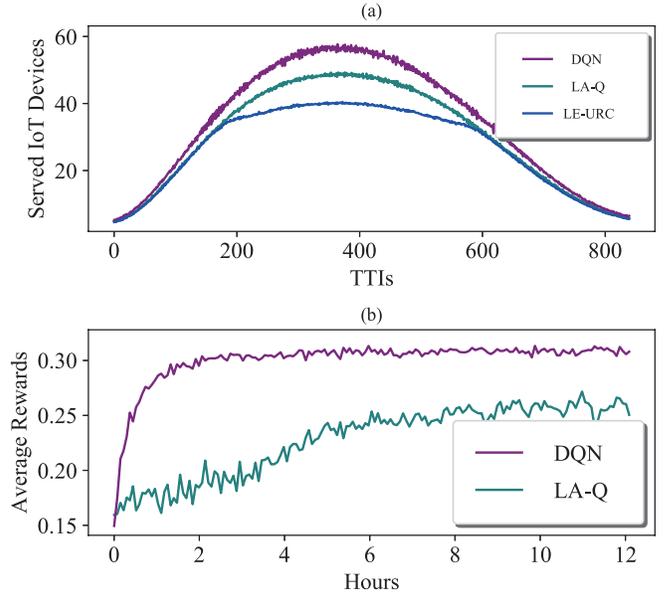}
        \vspace*{-0.2cm}
        \caption{\scriptsize The number of successfully served devices and the convergence speed.}
                \label{fig:6}
    \end{center}
    \vspace*{-0.4cm}
\end{figure}

\begin{figure*}[t]
    \centering
    \includegraphics[width=0.7\linewidth]{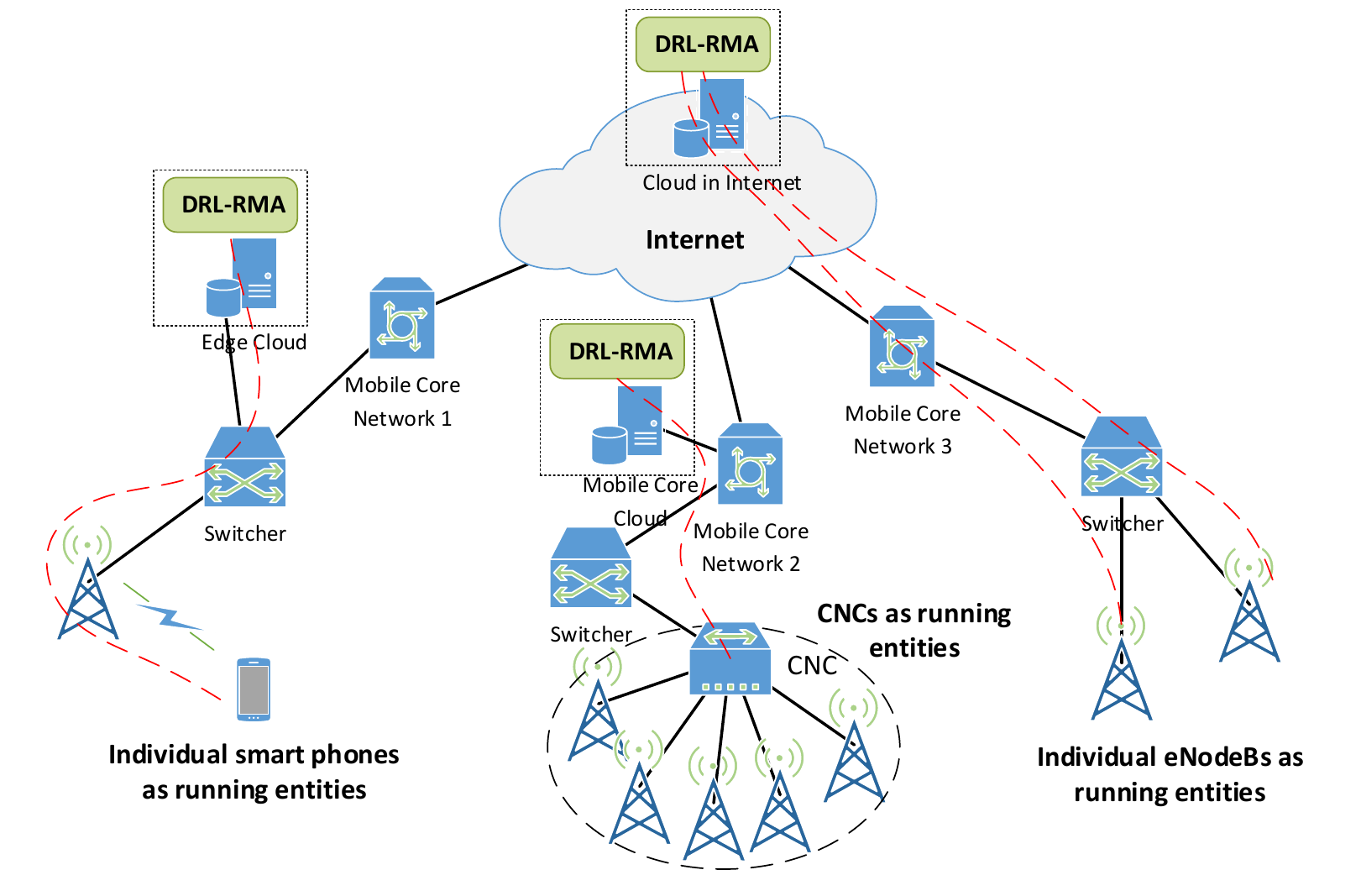}
    \caption{DRL based RRM providing ``DRL as a service" via: (1) service provider, (2) consumer entity, and (3) running entity. A third-party service provider runs and maintains DRL based resource management agent (DRL-RMA) on cloud, which could be either edge cloud in proximity of radio nodes, core cloud or remote cloud in Internet. }
    \label{fig:1}
\end{figure*}

\subsection{Case Study: RACH Access for Massive IoT}
In the following, a case study in random access is briefly illustrated. In cellular networks, the base station can observe the transmission receptions of both Random Access CHannel (RACH) and data transmission at the end of each time slot, which can be used to predict the traffic and facilitate the performance optimization of future time slots.  The complexity of the problem is compounded by the lack of a prior knowledge at the base station regarding the stochastic traffic and unobservable channel statistics. RL based on tabular-Q is not feasible for multi-parameter multi-group dynamic optimization in Narrow-Band IoT (NB-IoT) networks as shown in our work \cite{RL-IoT-MassC-JSAC} due to the large memory and high computation complexity required for the state-action value table, and the difficulty for the agent to repeatedly experience every state to achieve convergence within a limited time. 

This motivates the application of RL based on Linear Approximation (LA-Q) and DRL based on deep neural network (DQN) at the BS with guaranteed convergence capability within largely reduced training time. With the target of maximizing the long-term average number of devices that successfully transmit data, our results in Fig. \ref{fig:6} shown that both the DQN and LA-Q approaches outperform the conventional heuristic approaches based on load estimation (LE-URC) in current literature. More importantly, our proposed DQN approach outperforms our LA-Q approach in terms of the number of success devices  with much less training time.

The detailed mapping of DRL on RRM depends on specific problems and scenarios. A brief summary covering typical RRM issues is presented in Table \ref{table1}. As we have mentioned in Section I, the intensive computation complexity and energy consumption could hinder the application of DRL in future wireless networks. In addition, the features of RRM and networks are not fully investigated.

\section{Efficient DRL Architecture for RRM}
\label{section3}

To make DRL based RRM green, we envision a flexible cloud based DRL architecture for mobile networks. 

\subsection{Cloud based On-Demand DRL}
For traditional DL applications such as computer vision, DNN is trained and ran on centralized hardware resource (such as GPU and TPU). For RRM tasks, the running entities of bases stations or terminal devices could hardly afford the fee for sufficient computation resource deployment nor the associated energy consumption. Moreover, different from offline supervised training, training samples in RRM can be only generated from the interaction between RL agent and the environment and the instant reward feedback could be delayed and even could not be explicitly derived. For this reason, centralized and computation-intense DRL running architecture is not suitable for RRM. Instead, we proposed to decouple the DRL task and run it in a distributed and online manner to improve its efficiency and flexibility. 

In order to benefit various types of devices including those that can not afford the computing capability on his own, we envision a ``DRL as a service" approach by exploiting the benefits from cloud computing resources, as shown in Fig. \ref{fig:1}. Three roles are involved: (i) service provider, (ii) consumer entity, and (iii) running entity. A third-party service provider runs and maintains DRL based resource management agent (DRL-RMA) on cloud, which could be either edge cloud in proximity of radio nodes, core cloud or remote cloud in Internet. The service provider leases resource management service to different types of consumers by providing on-demand service via DRL-RMA. The introduction of virtualization to 5G has led to three different actors in networks: infrastructure provider, tenant, and the end-user. Infrastructure providers own and manage their physical networks and lease virtualized resources to tenants, and tenants offer network services to users using virtualized resources. Accordingly, infrastructure providers, tenants and users can be the DRL resource management consumers, depending on network deployments, business models (C2B, B2B), data-generation and processing pipelines, and demands. The running entities are devices that actually run RRM guided by DRL-RMA. For infrastructure providers and tenants, running entities can be a base station, CNC of dense cells, or even user equipments (UEs). While for users, running entities are their UEs, such as smart phones or tablets.

\subsection{Information Flow \& Learning Process}
The general information flow of DRL service can be described as follows. When a consumer sends a service request for resource management to the DRL-RMA, a DRL optimization process (DRL-OP) will be instantiated on the cloud with some necessary negotiation process. This process is responsible for ascertaining the requirements of the consumers, the problem mapping method, targeted running entities, DRL algorithm and other related parameters and requirements. After that, the DRL-RMA will allocate appropriate storage and computing resources and configure DRL algorithm for the DRL-OP. Finally, the DRL-OP will establish a connection with the specified running entities and guide their resource management. 
\begin{figure}[t]
    \centering
    \includegraphics[width=0.95\linewidth]{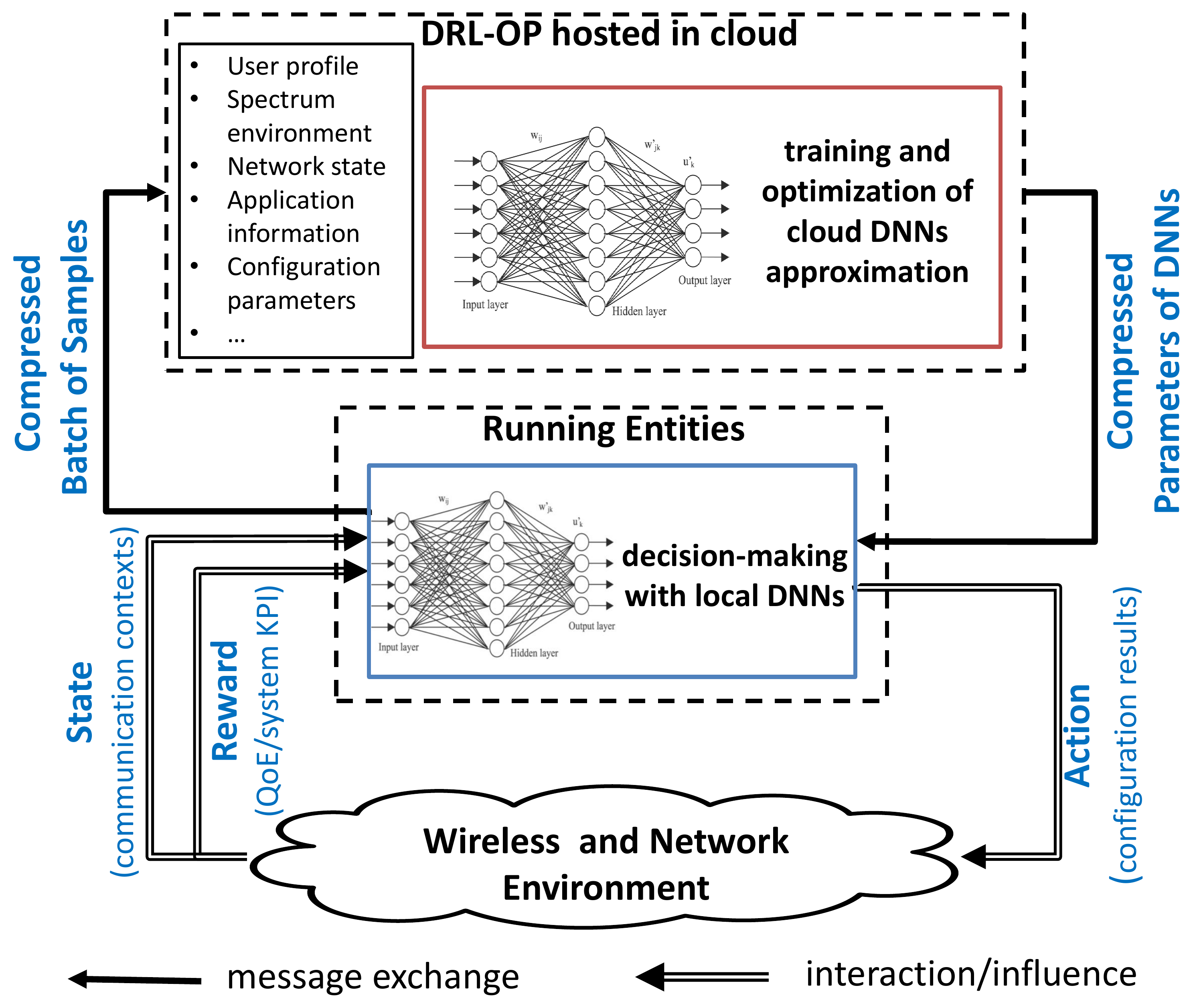}
    \caption{The interactive DRL loops.}
    \label{fig:2}
\end{figure}

Different from the agent-environment interactive loop of the DRL, the introduction of cloud results in two loops as shown in Fig. \ref{fig:2}. The inner loop is the running entity-environment interaction similar to that of DRL running locally. The additional outer loop is the message exchange between running entity and DRL-OP. Notably, as the training, optimization of neural networks and other computing-intensive operations are on the cloud, there is a special requirement on the optimization of DRL: mini-batch gradient descent rather than stochastic gradient descent is used to train deep neural networks. The reason is that stochastic gradient descent generally update gradient in a sample by sample manner, thus it may incur excessive message exchange cost between running entity and DRL-OP, especial for resource management problems with high decision frequency. On the contrary, mini-batch gradient descent allows for accumulating a set of samples for each update, which achieves a fair balance between message exchange cost and algorithm update frequency. More specifically, in each iteration of the outer loop, the DRL-OP sends a  copy of latest neural networks' parameters  (\eg, weight vectors) to the running entity; the running entity follows the recommended policy, \ie, updating a local neural network with the received parameters, and makes decisions with forward propagation computation for multiple interactions of inner loop; at the end of outer loop iteration, the running entity sends the accumulated samples (\ie, bath of samples) of ``state-action-reward" pairs to the DRL-OP for training. Obviously, one iteration of outer loop corresponds to multiple inner loop iterations.

In addition, different from DRL in computer games where learning samples could be easily generated, the samples are collected from the practical interaction between RRM entities and the wireless and network environment under the control of RL, which incurs more time and energy costs. This indicates that we actually have to employ incremental learning rather than one-shot training in DL. This is the reason why mini-batch of data samples are sent to the cloud periodically. To alleviate communication cost, both the sample batch and parameters of DNNs can be compressed to reduce the bandwidth requirement during the learning process.

This DRL as a service approach has several advantages. First, it provides flexible and on-demand resource management service for different consumers. Second, it offers devices with limited computation and battery capability a powerful optimization tool for parameter configuration. Third, it has limited influence on the network and traffic, as the service can be deployed in existing legacy cloud without additional infrastructure changes.

\section{Reducing DRL Complexity \& Energy via Algorithm Compression}
\label{section4}
In our previous section, we outlined an architecture that enables flexible on-demand RRM, but the computation complexity and energy consumption for DRL remain open challenges that are central to this paper. The size of DRL model largely determines the numbers of required operations and data access, which eventually determines the computation complexity and associated energy consumption. Thus, reducing the DRL algorithm size is crucial for cutting down computation complexity and energy consumption. In other words, we can compress the learning model to achieve a similar performance with significantly reduced parameters and energy consumption.

DRL can be treated as a combination of RL and DNN, where the former is responsible for the trade-off between exploitation and exploration in online RRM optimization, the latter is responsible for approximating policy or value functions for the former. The involving of RL loop in DNN indicates that the overall algorithm complexity is jointly determined by DNN and the underlying MDP in RL. Here, to achieve lightweight and energy-efficient DRL, we introduce to compress algorithm from three aspects: DNN model, MDP model and learning process.

\subsection{DNN Compression}
\begin{figure}[t]
    \centering
    \includegraphics[width=0.95\linewidth]{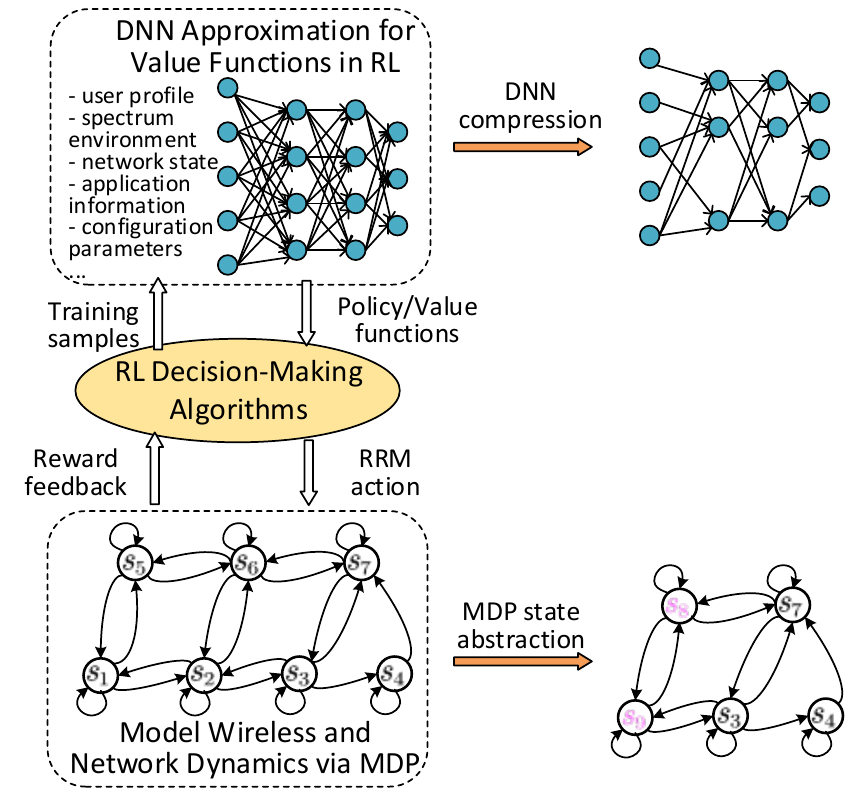}
    \caption{DNN and MDP compression.}
    \label{fig:3}
\end{figure}

The required operations and data access overhead in both training and inference of DNN are highly related with the numbers of neurons and the associated weights in it, \ie, a larger model size leads to higher energy consumption. Due to the lack of theoretical results on the optimal DNN architecture \footnote{Neuroevolution deep learning does offer a numerical pathway to finding optimal architectures}, current DNN in DRL application is generally designed based on experience, commonly resulting in a large model size and complexity than necessary. Nevertheless, previous studies have revealed that neural networks are typically over-parameterized, and there is significant redundancy that can be exploited \cite{sparsity}. Therefore, it is possible to achieve similar function approximation performance by removing redundant network architecture (pruning the network as shown in Fig. \ref{fig:3}) and only retaining useful parts with greatly reduced model size.  

There are several typical ways on compressing DNN by exploiting sparsity in neural networks. One method is reducing the number of parameters. This could be achieved by removing the number of connections/weights, \eg, weights smaller than some predefined threshold are removed, or pruning filters, \ie, removing redundant neurons and connections simultaneously. The second method is architectural innovations, such as replacing fully-connected layers with convolutional layers that is relatively more compact. Another method is weight quantization, \ie, reducing the precision of weights. For example, we can use 8-bit width integer rather than conventional 16-bit or 32-bit width floating-point number to store weights. Already, some of the aforementioned DNN compression practices have emerged in recent mobile deep learning applications.

\subsection{MDP Compression}
RL is commonly formulated under the framework of MDP or POMDP. The size of MDP is directly determined by the state and action spaces, which grow super-polynomially with the number of variables that characterize the domain. To support fine-grained RRM, we have to adopt high-resolution communication context to accommodate context-aware optimization, which often results in a large-scale MDP model. On the other hand, small model is always desired for improving energy-efficiency: a small state space will lessen the data storage space of samples and memory access cost in DNN training, and incur less exploration cost in terms of energy and time as sufficiently sampling the states is the intrinsic requirement of RL. Therefore, balancing the context characterization performance and model complexity is needed. For POMDP, in order to reduce the high dimensionality of the problem, hierarchical action space methods can be used to approximate the POMDP problem, achieving a scalable compression.

Considering that most DRL applications use discrete states and actions in DNN, we can compress MDP model in two stages. At the beginning of MDP modelling, we can appropriately choose the definitions of state and/or action to adjust their resolution. For example, when RSS is one dimension of state or transmit power constitutes action space, we could use a limit number of discretized levels to approximate their dynamic range with controlled performance loss. Besides, during the learning process, the size of MDP model can be further reduced by aggregating identical or similar states as shown in Fig. \ref{fig:3}, allowing us to reduce learning complexity with bounded loss of optimality \cite{state_abstraction}. The similarity of states can be measured in terms of optimal Q function, reward and state transitions, Boltzmann distributions on Q values and \etc. 

\begin{figure}[t]
    \centering
    \includegraphics[width=0.80\linewidth]{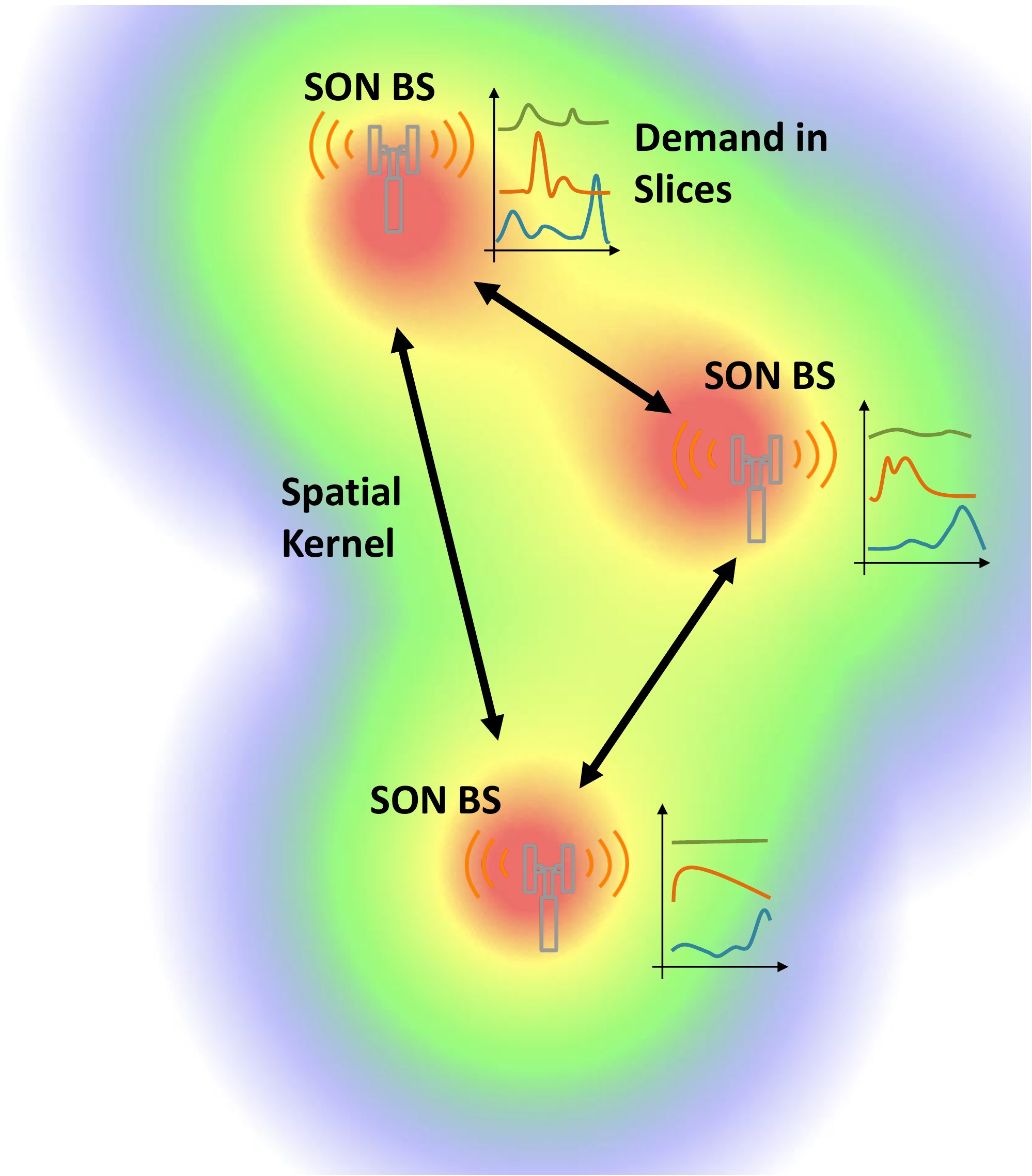}
    \caption{Spatial transfer learning among self-organization base stations.}
    \label{fig:4}
\end{figure}

\subsection{Improving Joint Efficiency via Spatial Transfer Learning}

Different from supervise learning tasks, the training samples in DRL can be only generated from the interaction between RL agent and the environment. As a result, accumulating training samples could be time-consuming. Accordingly, there will be a obvious delay before gathering sufficient samples to drive DNN. While existing DRL based RRM generally use simulated environment to generate training samples, they could hardly account for all effecting factors and the practical and complex dynamics of the system. One possible approach to compress the low-efficient learning process is transferring knowledge from similar RRM tasks. 

In cellular systems, hyper-dense deployment of base stations (BSs) will yield spatially correlated traffic demand patterns between neighbouring BSs. Here, we can exploit this phenomenon to achieve joint energy savings via spatial transfer learning. The widely used \textit{experience replay} is essentially transferring learning in time dimension, \ie, knowledge transfer from past learnt samples. Differently, the correlation among neighbouring BSs performing RRM tasks makes spatial transfer learning reasonable and novel.  

Many actions of multiple BSs need coordination, such as meeting user demand in a large event, offload traffic to each other, and sleep mode / cell expansion. Their RRM policies will have commonalities, which is represented by similar DNN parameters and/or RL policies. To exploit the spatial correlation, spatial transfer learning can be used between adjacent BSs (see example of potential application in densely deployed coordinating small cells with coordinated RRM \cite{DRL-HetNet}). We can use a spatial kernel that relates to the urban traffic correlation to transfer DRL function parameters among nearby wireless nodes. This is accomplished by first modelling dynamic spatial correlations between BSs using a flexible framework used commonly in disease and ecology modelling - stochastic integral-difference equation (SIDE), where a spatiotemporal dependent variable $z_{t}(s)$ evolves in accordance to $z_{t+1}(s) = \int_{O} k(s,r) f(z_{t}(r)) dr + e_{t}(s)$, over a spatial domain $s,r \in O$ and where $k(s,r)$ is the mixing kernel, $f(\cdot)$ is some function distorting the field and $e_{k}(s)$ is an added disturbance modeled as a Gaussian field. The SIDE links well with point process models of traffic demand, where a point process (Poisson or log-Gaussian Cox Process) has intensity $\lambda_{t}(s) \propto \exp{(z_{t}(s))}$. A SIDE function can be parameterized based on the traffic model correlation and then can be used to determine how much information to share between the DRL entities, by relating it to the correlation between traffic demand. The combined spatial DRL process can allow individual DRLs to learn faster by leveraging on the successful results of others, as shown in Fig. \ref{fig:4}. \\

\section{Open Challenges \& Conclusions}
\label{section5}

Here, we discuss pressing open challenges in achieving energy efficient DRL for RRM, inspiring the research community to act and collaborate.

\subsection{Lightweight DRL Training}
Current works on DNN compression mainly focus on reducing energy consumption in the inference stage of supervised learning algorithms but neglect the training stage that is more computation-intensive. In addition, in order to compress the inference process, the training process could be even more complex than the standard training without compression. The main reason is that many existing compression approaches rely on iterative training stage to discover the sparsity information and only reduce the DNN size progressively. Thus, reducing the complexity of DNN training stage is appealing. The challenge is due to the fact that DNN compression is problem specific. Unless the compressed DNN architecture information is given, we have to explore it during the training process.

\subsection{MDP Compression without Prior Information}
One drawback of the above mentioned MDP abstraction approaches is that they generally require to know the optimal solution of the MDP. This contradicts the motivation of using DRL to solve the underlying MDP in RRM context. Although the recent abstraction approach has relaxed this condition, the key parameters such as state transition and action value are needed, which is still impractical for the considered model-free RRM. Thus, how to abstract MDP states without prior information is a desiring and challenging task. One possible way is employing online abstraction, that is, as the progress of learning, we can reveal more information about the MDP model thus to abstract it progressively.

In conclusion, future AI driven automation of wireless networks and other critical infrastructures will bring about a step change in their ability to create efficient, resilient, and also user-centric services. However, the very same algorithms may also cause irreversible environmental damage due to their high energy consumption and lead to serious global sustainability issues. To achieve our goal of green AI for wireless networking, we have proposed several innovations and linked them to existing literature. On the running architecture, a cloud based on-demand DRL service model is proposed to provide computation capability and battery constrained devices with intelligent RRM. To reduce the computation and energy consumption in DRL, model compression approaches for reducing the sizes of DNN and MDP model are introduced. Finally, by exploiting the correlation feature of RRM tasks among nearby RRM entities, spatial transfer learning is further proposed to promote learning efficiency.

\bibliographystyle{IEEEtran}
\bibliography{RDLRRM}
\end{document}